# An Under-Explored Application for Explainable Multimodal Misogyny Detection in code-mixed Hindi-English

*Sargam Yadav[a], Abhishek Kaushik[b], Kevin Mc Daid[c]*

[a,b,c]Dundalk Institute of Technology
[a]sargam.yadav@dkit.ie
[b]abhishek.kaushik@dkit.ie (Corresponding author)
[c]kevin.mcdaid@dkit.ie

**Abstract**
Digital platforms have an ever-expanding user base, and act as a hub for communication, business, and connectivity. However, this has also allowed for the spread of hate speech and misogyny. Artificial intelligence models have emerged as an effective solution for countering online hate speech but are under explored for low resource and code-mixed languages and suffer from a lack of interpretability. Explainable Artificial Intelligence (XAI) can enhance transparency in the decisions of deep learning models, which is crucial for a sensitive domain such as hate speech detection. In this paper, we present a multi-modal and explainable web application for detecting misogyny in text and memes in code-mixed Hindi and English. The system leverages state-of-the-art transformer-based models that support multilingual and multimodal settings. For text-based misogyny identification, the system utilizes XLM-RoBERTa (XLM-R) and multilingual Bidirectional Encoder Representations from Transformers (mBERT) on a dataset of approximately 4,193 comments. For multimodal misogyny identification from memes, the system utilizes mBERT + EfficientNet, and mBERT + ResNET trained on a dataset of approximately 4,218 memes. It also provides feature importance scores using explainability techniques including Shapley Additive Values (SHAP) and Local Interpretable Model Agnostic Explanations (LIME). The application aims to serve as a tool for both researchers and content moderators, to promote further research in the field, combat gender based digital violence, and ensure a safe digital space. The system has been evaluated using human evaluators who provided their responses on Chatbot Usability Questionnaire (CUQ) and User Experience Questionnaire (UEQ) to determine overall usability.

**Keywords:** hate speech, misogyny, natural language processing, code-mixing, hinglish

## Introduction

Online hate speech and misogyny are alarmingly prevalent on social media platforms (Yadav, Kaushik, & Sharma, 2023) specially if we are dealing with multilingual society like India. (Rodriguez-Sánchez et al., 2020). In multilingual societies, users combine features of two or more languages in informal online communication, a phenomenon referred to as code-mixing (Mathur et al., 2018). For example, in Hinglish, Hindi is written in Romanized letters and combined with English words (Badjatiya et al., 2017). Existing research in hate speech detection focuses primarily on developing models for monolingual scenarios, or for specific modalities (Gomez et al., 2020) (Singh et al., 2024). Code-mixing further exacerbates the problem of hate speech detection using Artificial Intelligence (AI) models due to syntactic irregularities, inconsistent spelling and grammar, and noise (Bohra et al., 2018).Moreover, existing systems lack interpretability (Mathew et al., 2021), may suffer from unintended biases (Fersini et al., 2020), may be overly sensitive to the presence of curse words (Yadav, Kaushik, & McDaid, 2023), and sensitive to adversarial attacks (Gröndahl et al., 2018).

To address these challenges, we present an explainable multimodal misogyny detection system that supports text and images in code-mixed Hinglish and English[1]. The application utilizes transformer-based models such as multilingual Bidirectional Encoder Representations from Transformers (mBERT) (Devlin et al., 2019) and XLM-RoBERTa (XLM-R) (Conneau et al., 2019) for text-based misogyny identification, and ResNet (He et al., 2016) and EfficientNet (Tan & Le, 2019) combined with mBERT for multimodal classification. The application is also integrated with Shapley Additive Values (SHAP) (Lundberg & Lee, 2017) and Local Interpretable Model Agnostic Explanations (LIME) (Ribeiro et al., 2016) to provide feature attribution scores to enhance explainability and misogyny prediction. The application provides a user-friendly interface where images can be uploaded, selection can be performed between model and Explainable Artificial Intelligence (XAI) techniques, and confidence scores can be obtained for binary categorization as well as classification into subtypes of misogyny. The motivation behind conducting this study is to address the growing need to develop effective and inclusive systems for automatic hate speech detection. Lack of transparency in automatic hate speech detectors can negatively impact the free speech of target communities, highlighting the importance of XAI techniques.

---

[1] Demo video available at: demo.mp4

Additionally, we aim to address the gap in the availability of datasets in code-mixed languages, particularly for misogyny detection. The structure of the article is outlined below: Section 'Related Work' discusses the relevant literature in hate speech detection, explainability in hate speech detection, and evaluation of explainability techniques. Section 'Methodology' details the system architecture and methodology of the usability study, and Section 'Usability and Evaluation' discusses the results of usability explainability evaluation. Section 'Findings and Limitations' reports the findings and limitations of the study, and Section 'Conclusion' concludes the study.

## Related Work

Natural Language Processing (NLP) tools for hate speech detection have been studied extensively for languages such as English (MacAvaney et al., 2019), but the research in developing systems for under-resourced and code-mixed languages is nascent due to the lack of language specific tools, annotated datasets with the appropriate cultural context, and formal grammatical and linguistic rules (Mathur et al., 2018) (Santosh & Aravind, 2019). Several benchmark shared tasks including Hate Speech and Offensive Content Identification (HASOC) (Ranasinghe et al., 2023) and Automatic Misogyny Identification (AMI) (Fersini et al., 2020), are conducted regularly to promote further research and collaboration in countering hate speech detection (Yadav, 2024; Yadav, Kaushik, & McDaid, 2023). (Ranasinghe et al., 2022). Deep learning models generally perform better at hate speech classification, but suffer from a lack of transparency due to their black-box nature. XAI aims to develop tools to make the inner workings of opaque deep learning models more transparent and can help increase trust in model prediction. XAI techniques can be evaluated on dimensions such as faithfulness towards the model's prediction, stability to input perturbations, and plausibility to human evaluators (Narayanan et al., 2018). To evaluate the usability of web applications, usability studies are commonly conducted using standardized questionnaires to evaluate aspects such as the system's attractiveness, efficiency, and friendliness (Kaushik et al., 2021) (Kaushik & Jones, 2021, 2023a, 2023b; Sharma et al., 2021). In this study, the usability of the system will be evaluated using UEQ and CUQ. UEQ (Schrepp, 2015) is a well-established tool for assessing usability of a system across six dimensions. It consists of 26 items scored on a scale of -3 to +3. It also evaluates the hedonic and pragmatic quality of the system. CUQ is another effective tool to determine usability of a system or conversational agent (Ulster University, 2023). Participants respond to 16 questions, each rated from 1-5, with 1 indicating 'strongly agree' and 5 indicating 'strongly disagree'.

## Methodology

In this section, we will discuss details of the datasets used to train the models, model training and explainability, and the user interface of the system.

**System Architecture**

Ethical permission has been obtained from the host institute to collect data, perform weak annotation, and conduct the usability evaluation and recruit participants, ensuring compliance with institutional guidelines. The text-based misogyny detection models have been trained using a novel weakly annotated dataset in code-mixed Hinglish collected from Reddit, consisting of a total of 4,193 data points (title, body, and comments). In this work, weak annotation implies that the dataset was annotated by a single annotator, namely the primary author of this paper. Dataset annotation was performed at two levels: binary identification of misogyny (2,043 non-offensive and 2,150 misogynistic), and categorization into subtypes of misogyny into the following 10 labels: derogatory language (1,650), threats of violence/sexual violence fantasy (553), slut shaming (439), objectification and dehumanization (1,783), body shaming (101), victim blaming (87), stereotyping (373), minimization and trivialization (183), sexual harassment (10), and moral policing (7). For training multimodal models, we utilized the MIMIC dataset by (Singh et al., 2024), which includes images and memes annotated at two levels: binary identification of misogyny, and categorization into the following subtypes of misogyny: objectification, prejudice, and humiliation. Since we wanted to focus solely on code-mixed data, we have removed a total of 836 comments that were in Devanagari Hindi. The resulting dataset consisted of a total of 4,218 memes, consisting of 2,169 normal and 2,049 misogynistic memes, which were further divided into 1305 samples for 'objectification', 701 samples for 'prejudice', and 266 for 'humiliation'. Additionally, we have manually verified and changed the text extracted through Optical Character Recognition (OCR) extracted of approximately 849 samples to improve accuracy. OCR helps to automatically recognize characters and text that appears in images. For text-based misogyny identification, mBERT (binary macro F1-score: 0.93, multilabel macro F1-score: 0.27) and XLM-R (binary macro F1-score: 0.79, multilabel macro F1-score: 0.23) have been fine-tuned on the novel dataset and were selected due to their ability to effectively handle code-mixed Hinglish text. Hyperparameter tuning has been performed by varying the model's learning rate and training epochs, and the model has been evaluated using macro-F1 score. LIME and SHAP were implemented to provide feature attribution scores for a given input. For multimodal classification, an ensemble of EfficientNet + and mBERT (binary macro F1-score: 0.72, multilabel macro F1-score: 0.27), and ResNet and mBERT (binary macro F1-score: 0.74, multilabel macro F1-score: 0.32), was fine-tuned on the MIMIC dataset, following the model choices in (Singh et al., 2024). For multimodal classifiers, LIME and SHAP were implemented on textual and visual signals separately and produced a feature attribution plot for the textual input, and a saliency map for the visual input. The AI models have been implemented using PyTorch[2], LIME[3] and SHAP[4] have been implemented using their respective libraries, and Flask served as the web framework.

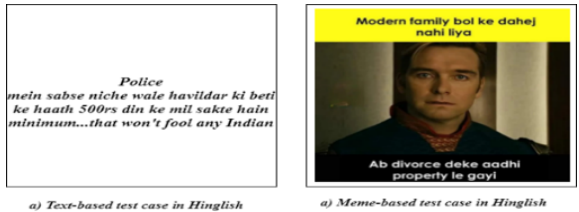

Figure 1: Sample Test Cases for Evaluation

**Usability Study**

To conduct usability studies, participants were recruited form the host institute and were made to sign an inform consent letter which informed them about the task and data privacy and anonymity. 6 participants were recruited, and were instructed to perform a task consisting of 3 test cases (either text or meme-based). Participants consisted of fellow PhD students in AI and its subfields, with varying degrees of familiarity with hate speech detection. They accordingly provided feedback on model prediction and explainability evaluation via the feedback forms in the application. After completion of these tasks, the users provided feedback on the overall system through UEQ and CUQ. Please note that the major focus of this study was to determine the usability of the misogyny classifier, and not focus on model performance, which will be explored in a future study. Figure 1 displays one text-based and one meme-based test case in Hinglish used to be perform usability evaluation. The Hinglish text in the text-based example 'Police mein sabse niche wale havildar ki beti ke haath 500rs din ke mil sakte hain minimum...that won't fool any Indian' translates to 'Even the daughter of the lowest-ranking officer in the police can easily get at least 500 rupees a day... that won't fool any Indian.'. The text in the meme 'Modern family bol ke dahej nahi liya Ab divorce deke aadhi property le gayi', translates to 'Called ourselves a modern family and didn't take dowry, she took half the property after divorce.'

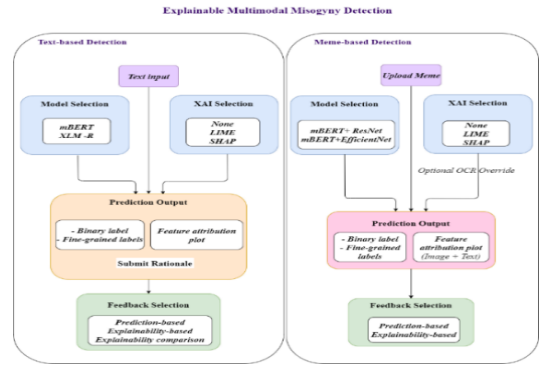

Figure 2: Flowchart of the System Architecture

Figure 2 displays the flowchart of the system, divided into two sections: text-based and meme-based. In the text-based detection section, the user inputs a sentence, selects a model for prediction (mBERT and XLM-RoBERTa), and submits the selection for analysis. The prediction results (binary and multi-label) are then displayed in the results section, which also includes an option to highlight rationale for prediction. The user can also select an XAI technique (LIME or SHAP), which will provide a feature attribution plot in the 'Results' section. Similarly, in the meme-based detection system, the user uploads a meme, selects a model (mBERT+ResNet or mBERT+EfficientNet) and an XAI technique (LIME or SHAP), and submits their selection to the system, which returns the binary and multi-label prediction scores, a feature importance plot for the textual signals, and a heatmap for visual explainability. Feedback can be provided on model prediction and explanations via the feedback forms. Figure 3 displays the prediction results with mBERT in the text-based section. The system predicts that the text is misogynistic with a confidence of 0.6104, and 'Objectification and dehumanization' as the fine-grained label with a confidence of 0.8339. Below the results, the figure displays the rationale selected by the user in multiple spans of text. The feedback section displays the prediction-based feedback form, which lists questions to determine the user's agreeability to the model's predictions.

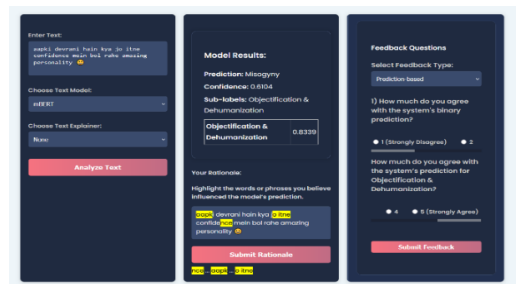

Figure 3: Snapshot of the system providing prediction and collecting rationale

---

[2] https://pytorch.org/
[3] https://github.com/marcotcr/lime
[4] https://shap.readthedocs.io/en/latest/

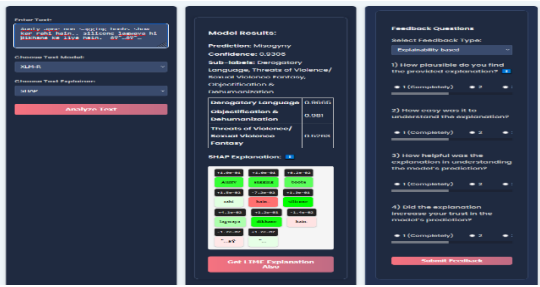

**Figure 4: Snapshot of the system providing prediction and explanation in the text-based section**

Figure 4 displays a screenshot of the system when XLM-R and SHAP were selected as the model and explainer. The model predicts that the sentence is misogynistic with a confidence of 0.9306, and 3 sublabels: Derogatory language (0.9665), Objectification and dehumanization (0.981), and Threats of violence/ sexual violence fantasy (0.5263). The result section also displays the feature importance scores using SHAP, which has highlighted tokens such as '*Aunty*' and '*sagging*' as being the biggest contributors to the prediction. The feedback section displays the explainability based feedback form with its respective questions.

## Usability and Evaluation

This section presents the results of the usability and explainability evaluations. Table 1 similarly reports the mean, standard deviation, and confidence interval for the six dimensions of UEQ. 'Stimulation' and 'novelty' have tight confidence intervals indicating consistent positive feedback. 'Dependability' has the highest variability (SD = 1.587) indicating inconsistent experience with the system.

**Table 1: UEQ Mean and Variance**

| Scale | Mean | Std. | CI |
|---|---|---|---|
| Attractiveness | 1.306 | 1.262 | [1.151, 1.460] |
| Perspicuity | 1.583 | 1.339 | [1.419, 1.748] |
| Efficiency | 1.125 | 1.367 | [0.957, 1.293] |
| Dependability | 0.625 | 1.587 | [0.430, 0.820] |
| Stimulation | 1.625 | 0.877 | [1.517, 1.733] |
| Novelty | 1.542 | 0.954 | [1.425, 1.659] |

The items of the CUQ questionnaire are summarized as follows with their means and standard deviations: Realistic and engaging (3.3 ± 1.0), Robotic (2.8 ± 0.8), Initially welcoming (2.7 ± 0.8), Unfriendly (2.2 ± 1.3), Scope and purpose explanation (2.7 ± 1.6), No indication of its purpose (2.8 ± 1.5), Easy to navigate (4.5 ± 0.8), Confusing (2.7 ± 1.6), Understanding (3.2 ± 1.0), Failed to recognize inputs (1.8 ± 1.3), Responses usefulness (4.0 ± 1.1), Irrelevant responses (2.0 ± 1.3), Coped well with errors (3.7 ± 1.2), Unable to handle errors (1.8 ± 1.0), Easy to use (4.2 ± 0.8), Complex (2.2 ± 1.5). The system achieved relatively high scores on several items such as ease of use, use of navigation, and useful and appropriate responses. The failure to recognize input and error handling were identified as critical weaknesses, both with a score of 1.8. Items 8 and 5 have highest standard deviations, indicating variation in user experience. UEQ scores on a scale of -3 to +3 and CUQ scores on a scale of 0 to 100. The minimum CUQ score is 42.2 and the maximum is 87.5, indicating a significant variation in usability experience. The mean score is 65.4, indicating that most users considered the system generally usable. A single score for UEQ was calculated by taking mean of the scores on the six dimensions for each user. The minimum score is -0.56 and the maximum is 2.88. Although most users report positive experiences (above 1.111), a significant outlier is the minimum score of -0.57, indicating a negative experience. Overall, the user-based evaluation of the system and explainability suggests that the system is considered moderately usable, with most users reporting a positive experience, and consistency in explanations.

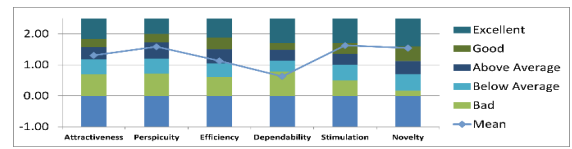

**Figure 5: UEQ Scores on Six Dimensions**

Figure 5 presents a breakdown of UEQ scores divided into six dimensions: attractiveness, perspicuity, efficiency, dependability, stimulation, and novelty. Bars show proportion of responses per category. The system has scored relatively high on 'stimulation' (~ 1.7), 'novelty' (~1.6), and 'perspicuity' (~1.5), indicating that the users perceived the system to be engaging, easy to understand, and simple to use. The weakest aspect of the system was 'dependability' with a score of 0.9, suggesting that users identified concerns about the reliability of the system. Wilcoxon signed-rank tests of UEQ means and CUQ means achieved scores of $p = 0.063$ and $0.375$, and therefore were not statistically significant at $\alpha = 0.05$. This could be attributed to the small sample size (n=6).

## Findings and Limitations

The findings of the usability study suggest that participants found the explainable multimodal misogyny classifier useful and intuitive, although the system struggled with dependability and efficiency. This could be due to delays in response times observed by the participants due to GPU-intensive AI models. This study has several limitations. The dataset used to train the text-based misogyny classifiers is relatively small and weakly annotated, and could therefore be biased. The participant size for the usability study was also small (n=6), and consisted of PhD students in AI-related areas.

# Conclusion and Future Work

This study introduced an explainable misogyny detection application that support multiple modalities and performed user-based evaluation to analyse the system's usability and the effectiveness of the explanations. The findings of UEQ and CUQ suggest that most users found the system easy to use and understand, but certain limitations such as lack of dependability and efficiency became prominent. The future scope of the study includes analysis of the feedback obtained from the prediction-based, explainability-based, and explainability comparison forms to focus on the quality of explanations.

## Acknowledgements
This research is partially funded through an internal fee waiver scholarship from Dundalk Institute of Technology (DkIT).